\documentclass[11pt]{article}

\usepackage[preprint]{acl}
\usepackage{times}
\usepackage{latexsym}

\usepackage[T1]{fontenc}

\usepackage[utf8]{inputenc}

\usepackage{microtype}

\usepackage{inconsolata}

\usepackage{graphicx}
\graphicspath{{figure/}}
\usepackage{amssymb}
\usepackage{amsmath}
\usepackage{tabularx}
\usepackage{array}
\usepackage{multirow}
\usepackage[most]{tcolorbox}
\usepackage{xcolor}

\usepackage{xspace} 
\newcommand{\method}{ReGuLaR\xspace}
\newcommand{\regformer}{ReGFormer\xspace}
\newcommand{\dataset}{\textsc{RGrounding-351K}\xspace}
\usepackage{booktabs}
\usepackage[table]{xcolor}

\newcommand{\gaincell}[1]{%
  \ifdim#1pt>0pt
    \textcolor{green!45!black}{\ensuremath{+#1}}%
  \else
    \ifdim#1pt<0pt
      \textcolor{red!75!black}{\ensuremath{#1}}%
    \else
      \ensuremath{#1}%
    \fi
  \fi
}

\title{ReGuLaR: Relation-Grounded Latent Reasoning for Large Vision-Language Models}

\author{
 \textbf{Zihu Wang\textsuperscript{1}},
 \textbf{Karthik Somayaji N.S\textsuperscript{1}},
 \textbf{Hejia Geng\textsuperscript{2}},
 \textbf{Peng Li\textsuperscript{1}}\\
 \textsuperscript{1}University of California, Santa Barbara\\
 \textsuperscript{2}University of Oxford\\
 \textsuperscript{1}\{zihu\_wang, karthi, lip\}@ucsb.edu
}

\begin{document}
\maketitle

\begin{abstract}
Chain-of-thought (CoT) reasoning has significantly improved the reasoning ability of large vision-language models (LVLMs) by verbalizing intermediate reasoning steps in natural language. However, such discrete textual rationales are often insufficient for encoding continuous visual evidence. Recent work addresses this limitation by moving reasoning into continuous latent space. Despite promising progress, existing methods leave latent reasoning insufficiently connected to the compositional and relational structure of visual evidence. To address this gap, we introduce \textbf{\method}, a relation-grounded latent reasoning framework that explicitly grounds latent states in this critical yet overlooked visual evidence. \method uses a training-time \textit{\regformer} to focus latent reasoning on question-relevant objects and inter-object relations, while at inference time the model reasons and generates answers without invoking the \regformer. To support training \method, we construct \dataset, a vision-language dataset annotated with key object bounding boxes and inter-object relations. Extensive experiments across diverse benchmarks show that \method consistently outperforms existing approaches and achieves state-of-the-art performance. 

\end{abstract}

\begin{figure}[t]
  \centering
  \includegraphics[width=0.97\columnwidth]{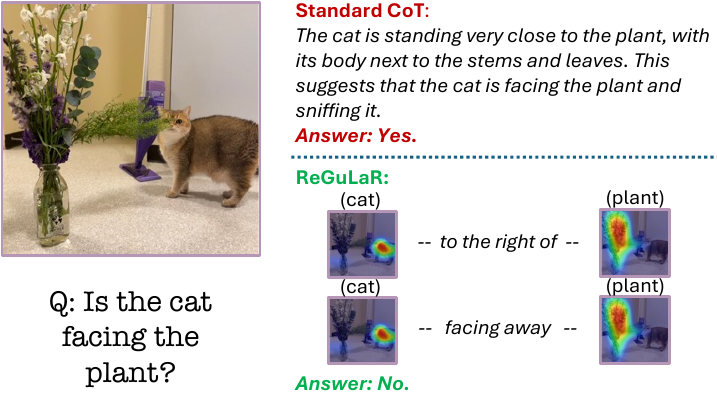}
  \caption{Unlike standard CoT, \method grounds latent reasoning in question-relevant objects and visual relations, forming a scene graph before generating the final answer.}
  \label{fig:intro}
\end{figure}

\begin{figure*}[t]
  \centering
  \includegraphics[width=0.91\linewidth]{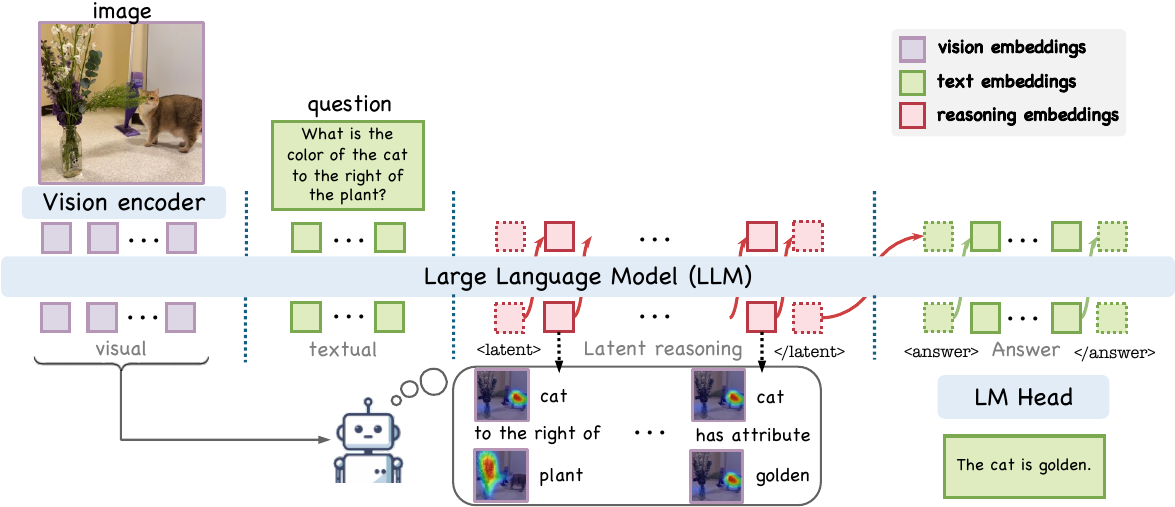} 
  \caption {Overview of \method. \method follows a thinking-then-answering process, where latent-space reasoning precedes final answer generation. At each latent reasoning step, the model focuses on one question-relevant object pair and their relation, or on one object and its attribute. A training-time \textit{\regformer} (Section~\ref{sec:regformer}) grounds each reasoning state in these critical visual structures, and is not required during inference.}
  \label{fig:overview}
\end{figure*}

\section{Introduction}

Recent large vision-language models (LVLMs) have shown impressive capabilities in jointly understanding visual and textual inputs~\citep{liu2023visual,bai2025qwen2,bai2025qwen3,achiam2023gpt}. Inspired by Chain-of-Thought (CoT) reasoning in language models~\citep{wei2022chain}, many LVLMs generate intermediate natural-language rationales before producing the final answer. However, discrete tokenized textual rationales are a lossy medium for representing continuous visual evidence, and they often provide only an implicit connection to the visual tokens that support each reasoning step~\cite{wang2026forest,li2025latent}. Recent latent reasoning approaches address this limitation by moving the reasoning process from natural language into continuous latent space. While representing rationales as latent embeddings is promising, existing methods primarily focus on latent attention optimization~\citep{jeon2026vision,ma2025multimodal,pham2025multimodal}, reconstructing visual information~\citep{li2025latent,tong2025sketch,yang2025machine}, optimizing latent reasoning trajectories~\citep{wang2026forest,liu2025reasoning,sun2025latent}, or interleaved vision-language reasoning~\citep{wang2025monet,chen2025reasoning,dong2025interleaved}. As a result, they often leave latent reasoning weakly connected to the compositional and relational structure of visual evidence. However, visual reasoning should not be defined only by where computation takes place, but also by the visual evidence around which it is organized. Images are composed of entities, attributes, and relations, and many visual questions require binding these elements into a question-relevant scene structure. Existing latent reasoning methods move reasoning into continuous hidden states, but provide limited control over whether the resulting latent rationales are grounded in this structured visual evidence.

To address this gap, we propose \textbf{\method}: \textbf{Re}lation-\textbf{G}ro\textbf{u}nded \textbf{La}tent \textbf{R}easoning for Large Vision-Language Models. Instead of exposing scene graphs as prompts or requiring graph generation at inference time, \method uses object relation-level supervision to shape question-conditioned latent states, inducing an internal graph-like reasoning process before answer generation. Representing graph-like visual evidence in the sequential model embeddings is challenging. As illustrated in Figure~\ref{fig:intro}, \method focuses on a question-relevant \texttt{subject-relation-object} relation triplet at each latent reasoning step, gradually forming a scene graph that captures the visual evidence needed to answer the question. Although appealing, encoding such an information-dense triplet into a single latent embedding remains challenging. To this end, we introduce a \textit{relation-grounding transformer} (\textit{\regformer}), which extracts subject and object visual features and grounds latent embeddings to the corresponding relations. To support the training of \method, we construct a high-quality image-text dataset, \dataset, with annotations of object relations and bounding boxes for the corresponding objects. Extensive experiments on diverse benchmarks show that \method consistently outperforms general purpose LVLMs, text-based CoT reasoning models and existing latent reasoning approaches, confirming \method achieves the state-of-the-art performance.

Our contributions are summarized as follows:
\begin{itemize}
    \item We propose \method, a relation-grounded latent reasoning framework that makes LVLMs reason over question-relevant objects and relations in continuous latent space before generating answers.

    \item We introduce \regformer, a training-time module that grounds latent reasoning states through role-aware subject/object attention and relation-level supervision, while preserving the standard inference interface without external scene graphs or object annotations.

    \item We construct \dataset, a vision-language dataset of approximately 351K examples, integrating question-answer pairs, scene graphs, and object bounding boxes to support fine-grained latent relation grounding.

    \item We demonstrate the effectiveness of \method on diverse visual reasoning benchmarks, where it outperforms strong general-purpose, RL-based, and latent reasoning baselines; ablations and visualizations further validate the importance and interpretability of relation-grounded latent supervision.
\end{itemize}

\section{Related Work}
\subsection{Structured Visual Evidence for Vision-Language Reasoning}

Large vision-language models (LVLMs) have evolved from task-specific architectures for applications like image captioning \citep{xu2015show,vinyals2015show} and visual question answering \citep{antol2015vqa,yang2016stacked} to large-scale pretrained models that align visual and textual representations \citep{radford2021learning,li2020oscar,alayrac2022flamingo}. With visual instruction tuning, recent LVLMs further connect strong vision encoders with large language models, showing strong capabilities in joint reasoning over visual and textual inputs \citep{liu2023visual,liu2024llavanext,bai2023qwen,bai2025qwen2,bai2025qwen3,wang2025vegas}.

Despite these advances, current LVLMs still struggle with fine-grained reasoning that requires faithful grounding of objects, attributes, spatial relations, and inter-object dependencies \citep{shiri2024empirical,fu2024blink}. Prior work introduces visual structures, such as localized object regions' bounding boxes and scene graphs, to provide more structured visual evidence \citep{chen2023shikra,wang2025llava,wang2026scenealign,huang2024structure}. However, it remains underexplored how to make answer generation follow a human-like process of first reasoning over question-relevant visual structures and then producing the answer. To this end, \method uses relation-level grounding to train a question-conditioned latent bottleneck, encouraging the model to organize relevant objects and relations into an implicit graph-like reasoning process before generating the answer.

\subsection{Latent Space Reasoning}
Chain-of-thought (CoT) improves reasoning by externalizing intermediate steps in natural language \citep{wei2022chain}, but discrete textual tokens can be verbose and lossy, especially for visual reasoning, where dense perceptual details are difficult to encode in words. Latent space reasoning addresses this limitation by allocating computation to hidden states rather than fully decoded rationales: prior work encodes complex intermediate information in the latent space of LLMs \citep{goyal2023think,zelikman2024quiet,xu2025softcot}, and more recent studies extend latent reasoning to multimodal models through latent attention optimization~\citep{jeon2026vision,ma2025multimodal,pham2025multimodal}, reconstructing visual information~\citep{li2025latent,tong2025sketch,yang2025machine}, optimizing latent reasoning trajectories~\citep{wang2026forest,liu2025reasoning,sun2025latent}, or interleaved vision-language reasoning~\citep{wang2025monet,chen2025reasoning,dong2025interleaved}. These methods show that reasoning can be more efficient and perceptually informative when performed in continuous space, but they do not explicitly ground the latent reasoning process in fine-grained object attributes and inter-object relations. In contrast, \method grounds latent reasoning with question-relevant object relations, encouraging the model to capture essential object attributes and relational dependencies during the reasoning process.

\section{Method}
\subsection{Overview}
As illustrated in Figure~\ref{fig:overview}, \method follows a human-like reasoning paradigm: it first performs latent reasoning and then generates the answer. Given an image-question pair, the model first enters the latent reasoning phase, where each reasoning step produces a latent state intended to encode one question-relevant object relation in the form of \texttt{subject-relation-object}. To guide these latent states toward such fine-grained visual information, we introduce a training-time \textit{relation-grounding transformer} (\textit{\regformer}), detailed in Section~\ref{sec:regformer}. At inference time, \method performs latent reasoning and answer generation without invoking \regformer. We describe the training data and optimization objective in Sections~\ref{sec:trainingdata} and~\ref{sec:trainingloss}, respectively.

\begin{figure}[t]
  \centering
  \includegraphics[width=0.95\columnwidth]{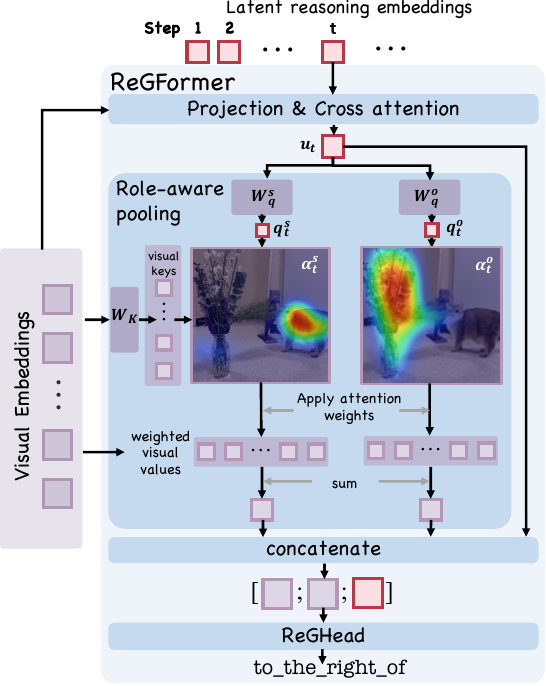}
  \caption{Illustration of \regformer. For each latent reasoning embedding, \regformer attends to visual tokens, performs role-aware pooling to extract subject and object evidence, and predicts the corresponding relation label.}
  \label{fig:regformer}
\end{figure}

\subsection{Relation Grounding with \regformer}
\label{sec:regformer}

Although scene graphs provide detailed and fine-grained visual information about an image, incorporating such graph-structured information into a sequential Chain-of-Thought is challenging. A straightforward approach is to verbalize the scene graph as natural language and include it in the reasoning trace. However, due to the discrete and sequential nature of text, this strategy can be token-expensive and lossy. Moreover, textual CoT is typically decoupled from the visual tokens that support each reasoning step, providing only an implicit connection between the rationale and the underlying visual evidence. While moving reasoning into continuous latent space is appealing, encoding complex graph-like visual information in latent states remains non-trivial.

To this end, we introduce \regformer, a training-time module that grounds latent reasoning states in fine-grained object relations and aligns them with the visual tokens that support each reasoning step. As illustrated in Figure~\ref{fig:regformer}, \regformer takes as input the latent reasoning state and the visual token embeddings from the backbone LVLM. For the $t$-th latent reasoning step, let $\mathbf{z}_t \in \mathbb{R}^{d}$ denote the hidden state of the current latent reasoning token, and let $\{\mathbf{v}_1,\dots,\mathbf{v}_n\}$ denote the $n$ visual token embeddings. We first project them into the relation-grounding space:
\begin{equation}
\mathbf{r}_t = \mathbf{z}_t \mathbf{W}_z, 
\qquad
\mathbf{x}_i = \mathbf{v}_i \mathbf{W}_v
\end{equation}
where $\mathbf{r}_t,\mathbf{x}_i \in \mathbb{R}^{d_r}$. The projected reasoning state attends to the visual tokens through $\mathrm{CrossAttn}$, a two-layer cross-attention block with gated residual fusion, producing $\mathbf{u}_t$:
\begin{equation}
\mathbf{u}_t = \mathrm{CrossAttn}(\mathbf{r}_t, \mathbf{X})
\end{equation}
Here, $\mathbf{X} = [\mathbf{x}_1;\dots;\mathbf{x}_n] \in \mathbb{R}^{n \times d_r}$.

Next, \regformer applies role-aware pooling to extract subject- and object-specific visual evidence for the current relation. We compute subject and object queries from $\mathbf{u}_t$ and visual keys from the projected visual tokens:
\begin{equation}
\mathbf{q}_t^s = \mathbf{u}_t \mathbf{W}_q^s,
\quad
\mathbf{q}_t^o = \mathbf{u}_t \mathbf{W}_q^o,
\quad
\mathbf{k}_i^v = \mathbf{x}_i \mathbf{W}_K
\end{equation}
where $\mathbf{q}_t^s,\mathbf{q}_t^o,\mathbf{k}_i^v \in \mathbb{R}^{d_r}$. We denote all visual keys as
$\mathbf{K}^v = [\mathbf{k}_1^v;\dots;\mathbf{k}_n^v] \in \mathbb{R}^{n \times d_r}$.
The subject and object attention distributions are then computed as:
\begin{equation}
\begin{aligned}
\boldsymbol{\alpha}_t^{s}
&=
\mathrm{softmax}
\left(
\frac{\mathbf{q}_t^{s}(\mathbf{K}^v)^\top}{\sqrt{d_r}}
\right) \\
\boldsymbol{\alpha}_t^{o}
&=
\mathrm{softmax}
\left(
\frac{\mathbf{q}_t^{o}(\mathbf{K}^v)^\top}{\sqrt{d_r}}
\right)
\end{aligned}
\end{equation}

Using these role-specific attention distributions, we obtain subject and object visual features by weighted pooling over the projected visual tokens:
\begin{equation}
\mathbf{c}_t^{s} = \sum_{i=1}^{n} \alpha_{t,i}^{s} \mathbf{x}_i,
\qquad
\mathbf{c}_t^{o} = \sum_{i=1}^{n} \alpha_{t,i}^{o} \mathbf{x}_i.
\end{equation}
Finally, we concatenate the subject feature, object feature, and relation-aware latent state:
\begin{equation}
\mathbf{m}_t = [\mathbf{c}_t^{s}; \mathbf{c}_t^{o}; \mathbf{u}_t]
\end{equation}
and feed $\mathbf{m}_t$ into a Relation-Grounding Head (ReGHead) to produce a distribution over the relation vocabulary. The resulting relation prediction and role-specific attention maps are used to supervise the latent reasoning state during training, while \regformer is removed during inference.

\subsection{Training Data}
\label{sec:trainingdata}

As described in Section~\ref{sec:regformer}, training \method with \regformer requires supervision for both object-level grounding and relation-level prediction. To this end, we construct \dataset, a high-quality image-text dataset with approximately 351K examples. Each example contains an image, a question-answer pair, a scene graph, and bounding boxes for the objects in the scene graph. For each image-question pair, we further derive a set of question-relevant relation targets from the image-level scene graph. Specifically, we identify objects mentioned in the question or answer as anchors using the available question/answer annotations, and select relations whose subject or object is connected to these anchors. Table~\ref{tab:dataset} summarizes the data sources of \dataset; construction, deduplication, and quality-control details are provided in Appendix~\ref{app:dataset}.

\begin{table}[t]
\centering
\small
\setlength{\tabcolsep}{3.5pt}
\renewcommand{\arraystretch}{1.05}
\begin{tabularx}{\columnwidth}{@{}>{\raggedright\arraybackslash}X r@{}}
\toprule
\textbf{Data Source} & \textbf{Amount} \\
\midrule
GQA~\citep{hudson2018gqa} & 245K \\
OpenImages~\cite{OpenImages2} & 50K \\
CLEVR~\citep{johnson2017clevr} & 25K\\
Visual Genome~\citep{krishna2017visual} & 15K\\
PSG~\citep{yang2022panoptic} & 6K\\
mGrounding~\citep{li2025migician} & 5K\\
VSR~\citep{liu2023visualsr} & 5K\\
\bottomrule
\end{tabularx}
\caption{Data composition of \dataset. We report the source datasets and the number of examples collected from each source.}
\label{tab:dataset}
\end{table}

\subsection{Training Objective}
\label{sec:trainingloss}
To train the model to follow the thinking-then-answering format, we apply supervised fine-tuning (SFT) with our proposed relation-grounding objectives. For each training example, we use the selected question-relevant relation targets in random order rather than the full scene graph. For each latent reasoning step $t$, we sample one subject-relation-object triplet from these targets, together with the bounding boxes of its subject and object. For attribute triplets, we use the bounding box of the described object for both subject and object grounding. We convert the bounding boxes into target attention distributions over visual tokens. Specifically, $\boldsymbol{\beta}_t^{s}, \boldsymbol{\beta}_t^{o}\in\mathbb{R}^{n}$ are defined as uniform distributions over the visual tokens whose image patches overlap with the subject and object bounding boxes, respectively, with zero probability assigned to all other tokens. Let $\boldsymbol{\alpha}_t^{s}$ and $\boldsymbol{\alpha}_t^{o}$ denote the subject and object attention distributions predicted by \regformer. We supervise role-aware grounding with:
\begin{equation}
\mathcal{L}_{\mathrm{attn}}
=
\frac{1}{T}
\sum_{t=1}^{T}
\left[
D_{\mathrm{KL}}\!\left(\boldsymbol{\beta}_t^{s} \,\|\, \boldsymbol{\alpha}_t^{s}\right)
+
D_{\mathrm{KL}}\!\left(\boldsymbol{\beta}_t^{o} \,\|\, \boldsymbol{\alpha}_t^{o}\right)
\right]
\end{equation}
where $D_{\mathrm{KL}}(\cdot\|\cdot)$ denotes the Kullback--Leibler (KL) divergence. Intuitively, this loss provides a grounding signal for the role-aware pooling layer in \regformer, encouraging its subject and object queries to attend to the corresponding visual evidence.

Given $\mathbf{m}_t$, ReGHead ($f_{\mathrm{rel}}$) produces logits over the relation vocabulary:
\begin{equation}
\boldsymbol{\ell}_t^{\mathrm{rel}}
=
f_{\mathrm{rel}}(\mathbf{m}_t),
\quad
\mathbf{p}_t^{\mathrm{rel}}
=
\mathrm{softmax}(\boldsymbol{\ell}_t^{\mathrm{rel}})
\end{equation}
We supervise relation prediction with the cross-entropy loss, encouraging each latent reasoning state to encode the relation between the currently grounded subject-object pair:
\begin{equation}
\mathcal{L}_{\mathrm{rel}}
=
-\frac{1}{T}
\sum_{t=1}^{T}
\log \mathbf{p}_{t,y_t}^{\mathrm{rel}}
\end{equation}
where $y_t$ is the ground-truth relation label for the $t$-th subject-object pair, and $\mathbf{p}_{t,y_t}^{\mathrm{rel}}$ denotes the probability assigned to the correct relation.

After the latent reasoning stage, the model generates the final answer autoregressively. We optimize the answer tokens with the standard next-token prediction loss:
\begin{equation}
\mathcal{L}_{\mathrm{ans}}
=
-\frac{1}{N}
\sum_{j=1}^{N}
\log
p_\theta
\left(
a_j
\mid
I, q, \mathbf{z}_{1:T}, a_{<j}
\right)
\end{equation}
where $I$ is the input image, $q$ is the question, $\mathbf{z}_{1:T}$ are the latent reasoning states, and $a_{1:N}$ are the answer tokens.

The overall training objective is:
\begin{equation}
\mathcal{L}
=
\lambda_{\mathrm{ans}}\mathcal{L}_{\mathrm{ans}}
+
\lambda_{\mathrm{rel}}\mathcal{L}_{\mathrm{rel}}
+
\lambda_{\mathrm{attn}}\mathcal{L}_{\mathrm{attn}}
\end{equation}

Through these supervision signals, \regformer guides the backbone LVLM to organize each latent reasoning step around a question-relevant \texttt{subject-relation-object} triplet before generating the answer. Because the relation prediction and attention grounding losses are computed from the latent states and back-propagated into the backbone LVLM, the model is encouraged to encode question-relevant subjects, objects, and relations in its own latent reasoning trajectory. Once this behavior is learned, \regformer is removed. At inference time, \method first performs a fixed number of latent reasoning steps and then switches to autoregressive answer generation, requiring only the image and question without scene graphs, object annotations, or additional grounding modules.

\begin{table*}[t]
\centering
\small
\setlength{\tabcolsep}{3pt}
\renewcommand{\arraystretch}{1.08}

\definecolor{opencolor}{RGB}{215,209,224}
\definecolor{closecolor}{RGB}{223,233,245}
\definecolor{greenbg}{RGB}{221,233,188}
\newcommand{\boldgaincell}[1]{{\boldmath\gaincell{#1}}}
\begin{tabular}{lcccccccc}
\toprule
\textbf{Models}
& \textbf{V*}
& \textbf{HRBench}
& \textbf{MMVP}
& \textbf{BLINK}
& \textbf{MMStar}
& \textbf{SEED-Bench-2-Plus}
& \textbf{HallusionBench}
& \textbf{Avg.}\\
\midrule

\rowcolor{greenbg}
\multicolumn{9}{c}{\textbf{\textit{General-purpose LVLMs}}} \\
GPT-4o
& 64.92 & 57.25 & 68.70 & 60.02 & - & - & - & - \\
LLaVA-OneVision
& 72.77 & 59.56 & 73.33 & 50.13 & 59.13 & 61.22 & 51.28 & 61.06 \\
Qwen2.5-VL-7B
& 76.44 & 64.88 & 70.33 & 57.02 & 59.67 & 65.31 & 57.04 & 64.38 \\
\midrule

\rowcolor{closecolor}
\multicolumn{9}{c}{\textbf{\textit{RL-based reasoning LVLMs}}} \\
DeepEyes
& 78.01 & 65.63 & 71.67 & 53.02 & 58.27 & 69.08 & 61.91 & 65.37 \\
PAPO
& 36.13 & - & 54.33 & 52.66 & 45.93 & 54.11 & 56.07 & - \\
Vision-R1
& 78.53 & 63.44 & 72.00 & 53.23 & 62.80 & 68.95 & 63.06 & 66.00 \\
\midrule

\rowcolor{opencolor}
\multicolumn{9}{c}{\textbf{\textit{Latent reasoning LVLMs}}} \\
LVR
& 79.06 & 50.63 & 70.33 & 54.97 & 58.13 & 47.39 & 65.99 & 60.93 \\
MONET
& 79.58 & 64.13 & 72.33 & 56.02 & 59.73 & 65.88 & 54.91 & 64.65 \\
LASER
& 80.10 & 65.82 & \underline{73.00} & \underline{58.55} & 61.20 & 70.05 & \underline{67.05} & 67.97 \\
SLVR-7B (Q1)$^{\dagger}$
& 82.20 & - & - & - & - & - & - & - \\
HyLaR-7B$^{\dagger}$
& \underline{83.77} & \textbf{72.75} & \textbf{73.67} & 57.14 & \underline{62.00} & \underline{70.32} & 63.68 & \underline{69.05} \\
\midrule

\textbf{\method-7B}
& \textbf{84.29} & \underline{69.94} & \textbf{73.67} & \textbf{61.49} & \textbf{62.47} & \textbf{71.01} & \textbf{67.14} & \textbf{70.00} \\
\addlinespace[2pt]
\rowcolor{black!5}
{\footnotesize\textit{vs Qwen2.5-VL-7B}}
& \boldgaincell{7.85} & \boldgaincell{5.06} & \boldgaincell{3.34} & \boldgaincell{4.47} & \boldgaincell{2.80} & \boldgaincell{5.70} & \boldgaincell{10.10} & \boldgaincell{5.62} \\
\bottomrule
\end{tabular}
\caption{Results on seven vision-language reasoning benchmarks. Scores are accuracy (\%). $\dagger$ marks author-reported results. Avg. requires all seven benchmark overall scores. \textbf{Bold} and \underline{underlined} numbers denote the best and second-best scores among latent reasoning LVLMs in each column, respectively.}
\label{tab:main_results}
\end{table*}

\section{Experiment}

\paragraph{Experiment setup.} 
We use Qwen2.5-VL-7B~\citep{bai2025qwen2} as the backbone and initialize \method from the corresponding pretrained checkpoints. During training, we freeze the vision encoder and the modality-alignment projector. We set the loss weights $\lambda_{ans}=1.0$, $\lambda_{rel}=1.0$, and $\lambda_{attn}=0.1$ for our main experiments. Training is conducted on $4\times$ NVIDIA A100 GPUs with 80GB memory. Inference latency is profiled on a single NVIDIA A100-80GB PCIe GPU with batch size 1. Unless otherwise specified, we use eight latent reasoning steps at inference. Additional implementation details are provided in the Appendix.

\paragraph{Evaluation benchmarks.}
We evaluate the reasoning ability of \method on a diverse set of vision-language benchmarks with English questions and answers. 
\textbf{$V^*$ Bench}~\citep{wu2024v} evaluates a model's ability to recognize fine-grained visual details and relative spatial relations. 
\textbf{MMVP}~\citep{tong2024eyes} evaluates fine-grained visual perception by testing whether LVLMs can distinguish visually similar images and answer questions about details that are often overlooked.
\textbf{HRBench}~\citep{wang2025divide} assesses model performance on reasoning over ultra-high-resolution images. 
\textbf{BLINK}~\citep{fu2024blink} is a comprehensive benchmark covering 14 visual perception tasks. 
\textbf{MMStar}~\citep{chen2024we} evaluates visual perception and multimodal reasoning with image-dependent questions spanning six core capabilities.
\textbf{SEED-Bench-2-Plus}~\citep{li2024seed} evaluates text-rich visual comprehension across diverse real-world scenarios such as charts, maps, and web pages.
\textbf{HallusionBench}~\citep{guan2024hallusionbench} tests whether LVLMs can avoid hallucinations and visual illusions under subtle visual changes.

\paragraph{Baselines.}
We compare \method with models from three categories. 
(1) General-purpose LVLMs, including LLaVA-OneVision~\citep{li2024llava}, GPT-4o~\citep{achiam2023gpt}, and Qwen2.5-VL-7B~\citep{bai2025qwen2}. 
(2) RL-based reasoning LVLMs, including DeepEyes~\citep{zheng2025deepeyes}, PAPO~\citep{wang2025perception}, and Vision-R1~\citep{huang2025vision}. 
(3) Latent-reasoning LVLMs, including LVR~\citep{li2025latent}, MONET~\citep{wang2025monet}, LASER~\citep{wang2026forest}, SLVR~\citep{xu2026semantic}, and HyLaR~\citep{cheng2026hylar}.
We use author-reported results for SLVR and HyLaR. Baseline backbones and locally evaluated checkpoints are listed in Appendix~\ref{app:checkpoints}.

\subsection{Main Results}
As shown in Table~\ref{tab:main_results}, \method achieves the best accuracy on six of seven benchmarks among the compared latent reasoning methods, including a tie on MMVP, and the highest average accuracy.

\method performs strongly across $V^*$ Bench, MMVP, HRBench, BLINK, and MMStar, improving over its pretrained backbone by 2.80--7.85 percentage points. In particular, it reaches 84.29\% on $V^*$ Bench and 61.49\% on BLINK, where fine-grained attributes and object relations are central. These gains suggest that supervising latent states with relation labels and localized subject--object evidence helps preserve visual details and organize them for reasoning.

Beyond general perception, \method also leads the compared latent reasoning methods on MMStar (62.47\%), which includes science and mathematics, and SEED-Bench-2-Plus (71.01\%), which covers text-rich charts, maps, and web pages. 
On HallusionBench, \method reaches 67.14\% accuracy, exceeding its backbone by 10.10 percentage points and supporting improved robustness to hallucinations and visual illusions. Together, these broader results suggest that organizing question-relevant evidence in grounded latent states supports reasoning over diverse visual content while helping constrain predictions unsupported by the image.

Benchmark breakdowns are provided in Appendix~\ref{app:category_results}, and limited-data training results in Appendix~\ref{app:data_efficiency}.

\begin{figure}[t]
  \centering
  \includegraphics[width=\columnwidth]{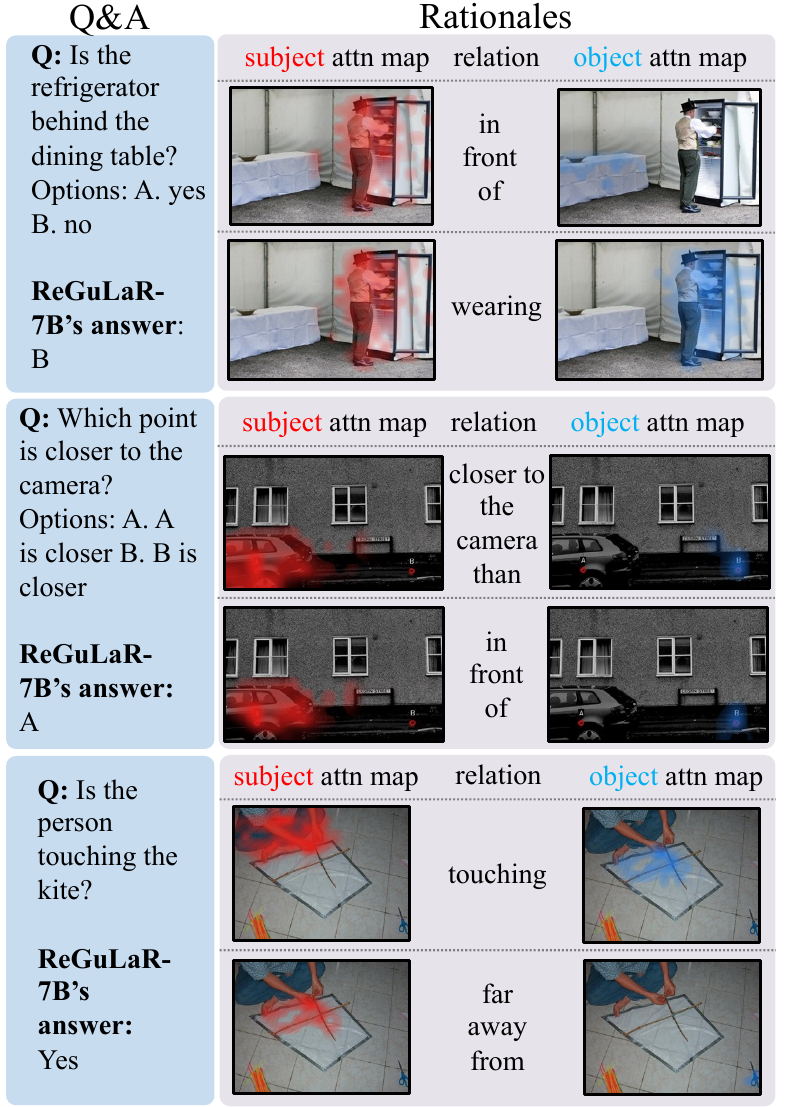} 
  \caption{Visualization of relation-grounded latent rationales produced by \method. For each example, we show the input question, the model prediction, and two latent reasoning steps decoded by the trained \regformer. In each step, the red heatmap indicates the subject attention, the blue heatmap indicates the object attention, and the middle text shows the relation predicted from the corresponding latent reasoning state.}
  \label{fig:visualization}
\end{figure}

\subsection{Visualization of Relation-Grounded Latent Rationales}
\label{sec:visualization}

Although \regformer is not required at inference time, the trained \regformer can still be attached to the latent reasoning states as a diagnostic probe to visualize their semantics. Figure~\ref{fig:visualization} shows representative examples, each including the input question, the answer of \method, and two latent reasoning steps. For each step, we visualize the subject attention map in red, the object attention map in blue, and the relation predicted by \regformer from the corresponding latent state. These examples show that different latent states can focus on question-relevant entities and recover meaningful relations, suggesting that \method performs visually grounded relation reasoning before final answer generation.

\subsection{Ablation Study}

\paragraph{Effect of Relation-Grounded Latent Supervision.}
To further demonstrate the effectiveness of our proposed relation-grounded latent reasoning, we introduce two additional models fine-tuned from Qwen2.5-VL-7B on \dataset. \textit{Vanilla SFT} is trained only with the next-token prediction loss to directly generate the answer without latent reasoning. The \textit{Text Rationales} variant converts the question-relevant scene graph into textual subject-relation-object triplets and uses them as intermediate rationales before answer generation. As shown in Table~\ref{tab:ablation_rationale}, \method substantially outperforms Vanilla SFT, indicating that the improvement does not simply come from fine-tuning on \dataset. Instead, explicitly modeling a thinking-then-answering process helps the model organize the visual evidence needed for reasoning. Moreover, \method also outperforms the Text Rationales variant, showing that verbalizing visual relations as discrete text is less effective than grounding continuous latent states with visual-token evidence. This suggests that \method benefits from both the compactness of continuous latent reasoning and the fine-grained evidence support provided by \regformer over visual tokens.

\begin{table}[t]
\centering
\small
\setlength{\tabcolsep}{2mm}
\renewcommand{\arraystretch}{1.08}
\begin{tabular}{lcccc}
\toprule
\textbf{Method} & \textbf{V*} & \textbf{MMVP} & \textbf{BLINK} & \textbf{Avg.} \\
\midrule
Qwen2.5-VL-7B 
& 76.44 & 70.33 & 57.02 & 67.93 \\
Vanilla SFT 
& 73.82 & 71.00 & 56.50 & 67.11 \\
Text Rationales 
& 74.87 & 70.33 & 57.12 & 67.44 \\
\method 
& \textbf{84.29} & \textbf{73.67} & \textbf{61.49} & \textbf{73.15} \\
\bottomrule
\end{tabular}
\caption{Ablation study on relation-grounded latent supervision. We compare \method with the base Qwen2.5-VL-7B, Vanilla SFT on the same training data, and the Text Rationales variant that verbalizes visual relations in the reasoning trace.}
\label{tab:ablation_rationale}
\end{table}

\begin{figure}[t]
\centering
\includegraphics[width=0.96\columnwidth]{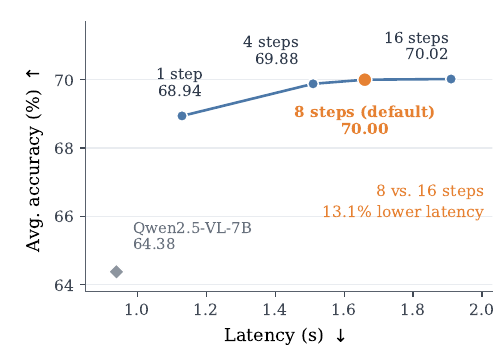}
\caption{Accuracy and latency across latent reasoning budgets. Accuracy is the mean of accuracies across the seven benchmarks in Table~\ref{tab:main_results}; latency is mean end-to-end inference time.}
\label{fig:reasoning_efficiency}
\end{figure}

\paragraph{Number of reasoning steps.}
At inference time, \method uses a fixed budget of latent reasoning steps before generating the final answer. We study the accuracy--latency trade-off across 1, 4, 8, and 16 steps in Figure~\ref{fig:reasoning_efficiency}, reporting mean overall accuracy across the seven benchmarks in Table~\ref{tab:main_results} and mean end-to-end latency on $V^*$ Bench. Increasing the budget from one to eight steps improves average accuracy from 68.94\% to 70.00\%. Doubling the budget to sixteen steps yields only a 0.02 percentage-point gain, while latency increases from 1.66 to 1.91 seconds. We therefore use eight latent reasoning steps by default, reducing latency by 13.1\% relative to sixteen steps while maintaining comparable average accuracy.

\paragraph{\regformer architecture.}
We examine how \regformer's architecture affects downstream visual reasoning. 
The full module concatenates subject/object visual features and the visually grounded latent state to form the input to ReGHead, $[\mathbf{c}_t^s;\mathbf{c}_t^o;\mathbf{u}_t]$. \textit{w/o $\mathbf{u}_t$ in ReGHead} removes $\mathbf{u}_t$ from this concatenation, using only $[\mathbf{c}_t^s;\mathbf{c}_t^o]$. 
\textit{Global mean pooling} replaces the two role-specific features with the mean of all projected visual tokens, yielding $[\bar{\mathbf{x}};\mathbf{u}_t]$. 
Table~\ref{tab:ablation_regformer} shows that average accuracy across V*, MMVP, and BLINK drops for both variants. 
These results support \regformer's use of both localized object features and latent-conditioned scene context: object localization alone may not determine the relation, while global pooling does not explicitly indicate which relation triplet to focus on at each step.

\begin{table}[htbp]
\centering
\small
\setlength{\tabcolsep}{2pt}
\renewcommand{\arraystretch}{1.08}
\begin{tabular}{@{}lcccc@{}}
\toprule
\textbf{Variant} & \textbf{V*} & \textbf{MMVP} & \textbf{BLINK} & \textbf{Avg.} \\
\midrule
Qwen2.5-VL-3B & 72.25 & 60.33 & 53.02 & 61.87 \\
w/o $\mathbf{u}_t$ in ReGHead & 71.73 & 60.33 & 52.81 & 61.62 \\
Global mean pooling & 72.25 & 61.00 & 52.55 & 61.93 \\
\midrule
\method-3B & \textbf{74.35} & \textbf{62.00} & \textbf{55.13} & \textbf{63.83} \\
\bottomrule
\end{tabular}
\caption{\regformer architecture ablation with Qwen2.5-VL-3B. Trained variants use the same 10K training subset and 1,000 optimization steps. Avg. is the mean accuracy (\%) on V*, MMVP, and BLINK.}
\label{tab:ablation_regformer}
\end{table}

\paragraph{Relation triplet ordering and relevance.}
GQA \citep{hudson2018gqa} annotates each question with a semantic program: an ordered list of operations needed to derive the answer, such as locating objects, following relations between them, and checking their attributes. We order question-relevant triplets according to the sequence of relation operations in the program and compare this supervision order with a fixed random permutation of the same triplets. To test relevance, we use count-matched question-irrelevant triplets from the same image, keeping other training settings fixed. Table~\ref{tab:ablation_triplet_order} shows similar average accuracies for program and random ordering. Irrelevant triplets lower accuracy on all three benchmarks and reduce the average to 60.50\%, 2.53 percentage points below random ordering. These results suggest that triplet relevance matters more than ordering in this setting; we therefore use random triplet ordering by default.

\begin{table}[htbp]
\centering
\small
\setlength{\tabcolsep}{2pt}
\renewcommand{\arraystretch}{1.08}
\begin{tabular}{@{}lcccc@{}}
\toprule
\textbf{Targets} & \textbf{V*} & \textbf{MMVP} & \textbf{BLINK} & \textbf{Avg.} \\
\midrule
Semantic-program order & 72.77 & \textbf{61.67} & 54.02 & 62.82 \\
Fixed random order & \textbf{73.30} & \textbf{61.67} & \textbf{54.13} & \textbf{63.03} \\
Question-irrelevant & 69.63 & 58.33 & 53.55 & 60.50 \\
\bottomrule
\end{tabular}
\caption{Relation triplet ordering and relevance on 5K GQA examples, using Qwen2.5-VL-3B trained for 1,000 optimization steps. The first two rows use the same question-relevant triplets; the last uses matched question-irrelevant triplets from the same image. Avg. is the mean accuracy (\%) on V*, MMVP, and BLINK.}
\label{tab:ablation_triplet_order}
\end{table}

\section{Discussion}

In this work, we present \method, a relation-grounded latent reasoning framework for LVLMs. Instead of verbalizing intermediate reasoning steps as natural-language CoT or leaving latent thoughts weakly grounded, \method uses question-relevant scene graphs to supervise latent reasoning states during training. 
With the training-time \regformer, each latent state is encouraged to encode the subject, object, and relation information needed for faithful visual reasoning, while inference remains simple and requires no external scene graphs, object annotations, or additional grounding modules. 
Extensive experiments show that \method consistently improves over strong existing approaches across diverse benchmarks. 
Future work may extend this framework to dynamic scenes by modeling temporal relations, object interactions, and state changes.

\section{Limitations}

One limitation of \method is its reliance on relation-level grounding supervision during training, including object bounding boxes and inter-object relation annotations. Although these annotations are not needed at inference time, collecting high-quality relation supervision may become expensive when scaling to broader domains. Nevertheless, the limited-data ablation in Appendix~\ref{app:data_efficiency} shows useful improvements over the pretrained backbone with small annotated training pools. Future work may reduce this requirement through weak supervision, automatic scene-graph construction, or self-training pipelines that produce relation signals at scale.

\bibliography{acl_latex}

\clearpage
\appendix

\section{Implementation Details}
\label{app:training_details}

\subsection{Training setup}
We initialize \method from the pretrained Qwen2.5-VL-7B checkpoint. The vision encoder and the modality-alignment projector are frozen during training, while the language model and the relation-grounded latent reasoning components are updated. The model is trained on \dataset with supervised fine-tuning and the proposed relation-grounding objectives.

We train \method for 3000 optimization steps on $4\times$ NVIDIA A100 GPUs with 80GB memory. We use a per-device batch size of 1 and gradient accumulation of 16, resulting in a global batch size of 64. The base learning rate is $1\times10^{-5}$ with a cosine learning-rate schedule and a warmup ratio of 0.03. We use AdamW with $\epsilon=10^{-8}$ and weight decay of 0.1. Training is performed in bfloat16 with gradient checkpointing and DeepSpeed ZeRO-3. For the training objective, we set $\lambda_{\mathrm{ans}}=1.0$, $\lambda_{\mathrm{rel}}=1.0$, and $\lambda_{\mathrm{attn}}=0.1$. The training process takes approximately 70 hours to complete.

Unless otherwise specified, all results are obtained from a single training run and evaluated with deterministic decoding, following the practice of recent latent reasoning work~\citep{li2025latent,wang2026forest,wang2025monet}. The Avg. column reports the arithmetic mean of benchmark-level overall accuracies.

We use dynamic image resolution. Each image is constrained to 128 to 8,192 visual tokens, corresponding to an image-size budget from $128 \times 28 \times 28$ to $8192 \times 28 \times 28$ pixels. 

\paragraph{Latent-state generation.}
The reasoning phase starts with \texttt{<|regular\_start|>}, whose final-layer hidden state initializes the first latent step. Subsequent steps directly use the preceding hidden state as their input embedding, without decoding it into text.

\paragraph{Training length and supervision.}
Training uses one slot per retained triplet in a random order fixed across updates. Each triplet supplies its slot's relation label and subject/object-box supervision. Ordinary examples retain all selected targets; full-graph fallback examples are capped at 16, using question/answer lexical matches followed by deterministic random selection. Shorter examples use fewer slots; batch padding is masked out.

\paragraph{Inference budget.}
Inference requires neither \regformer nor relation annotations. We allow at most eight reasoning steps by default. An earlier \texttt{<|regular\_end|>} ends reasoning, while reaching the budget forces this token. Answer generation then follows; unused reasoning steps are not padded.

\subsection{\regformer Architecture}
\label{app:regformer_architecture}

\regformer is a lightweight training-time module built on top of Qwen2.5-VL-7B. The backbone hidden size is $d=3584$, and both latent reasoning states and visual token states are projected into a relation-grounding space of dimension $d_r=1024$. We use two linear projections for this mapping, one for latent reasoning states and one for visual token states.

The projected states are passed through a 2-layer cross-attention transformer with 8 attention heads. In each layer, latent reasoning states serve as queries, while visual token states serve as keys and values. Each layer contains LayerNorm, multi-head cross-attention, a residual connection, and a feed-forward network with hidden dimension $4d_r=4096$ and GELU activation. The dropout rate is set to 0.0. The transformed reasoning state is combined with the original projected reasoning state through a learnable residual gate initialized to $10^{-3}$.

For role-aware pooling, \regformer uses separate $1024\times1024$ linear projections to generate subject queries, object queries, and visual keys. The subject and object queries attend to all visual tokens separately, producing role-specific attention maps and visual features. The subject feature, object feature, and relation-aware latent state are concatenated into a 3072-dimensional vector, normalized with LayerNorm, and passed to ReGHead, a two-layer MLP. ReGHead maps 3072 dimensions to 1024 with GELU activation, and then outputs logits over the final relation vocabulary of 2,111 labels. \regformer is used only during training and removed during inference.

\subsection{Ablation Baseline Training Details}
\label{app:ablation_baselines}

In the ablation study, we compare \method with two baselines: \textit{Vanilla SFT} and \textit{Text Rationales}. Both baselines are initialized from the same Qwen2.5-VL-7B checkpoint and trained on \dataset. To ensure a fair comparison, we keep the main training setup consistent with \method, including the frozen vision encoder and modality-alignment projector, optimizer, learning-rate schedule, batch size, number of training steps, and image resolution budget.

\paragraph{Vanilla SFT.}
The Vanilla SFT baseline is trained to directly generate the final answer without latent reasoning. It uses the same image-question-answer examples from \dataset, but removes the latent reasoning tokens, scene-graph triplet supervision, and \regformer objectives. The model is optimized only with the standard next-token prediction loss over the answer tokens.

\paragraph{Text Rationales.}
The Text Rationales baseline keeps the thinking-then-answering format but represents intermediate reasoning as natural language. Specifically, we convert the question-relevant scene-graph annotations into textual subject-relation-object triplets and insert them before the final answer as explicit rationales. For example, the textual rationale is formatted as a sequence of triplets such as \texttt{refrigerator-in\_front\_of-dining table; person-wearing-shirt}, followed by the final answer. The model is trained with next-token prediction over both the textual triplet rationales and the final answer. This baseline uses the same data and training recipe as \method, but replaces relation-grounded latent supervision with explicit text-based rationale supervision.

\subsection{Loss-Weight Sensitivity}
\label{app:loss_weights}
We study sensitivity to the relation prediction loss weight $\lambda_{\mathrm{rel}}$ and the attention grounding loss weight $\lambda_{\mathrm{attn}}$. We vary one weight at a time, fixing $\lambda_{\mathrm{ans}}=1.0$ and the other auxiliary weight at its default value ($\lambda_{\mathrm{rel}}=1.0$, $\lambda_{\mathrm{attn}}=0.1$). All variants use eight latent reasoning steps at inference. Table~\ref{tab:ablation_lambda} reports mean accuracy over $V^*$ Bench, MMVP, and BLINK. The default configuration achieves the highest mean accuracy of 73.15\%, matching the full-model result in Table~\ref{tab:ablation_rationale}. Within the tested ranges, both smaller and larger auxiliary weights reduce accuracy, supporting a balance between relation-grounding supervision and answer generation.

Notably, removing relation supervision ($\lambda_{\mathrm{rel}}=0$) lowers mean accuracy to 65.44\%, while removing bounding-box attention supervision ($\lambda_{\mathrm{attn}}=0$) lowers it to 66.79\%, corresponding to drops of 7.71 and 6.36 percentage points, respectively. Both variants retain the same reasoning architecture and answer supervision, with the remaining auxiliary loss at its default weight. These controlled ablations support the contribution of each grounding objective within our latent reasoning framework.

\begin{table}[htbp]
\centering
\small
\setlength{\tabcolsep}{1.7mm}
\renewcommand{\arraystretch}{1.08}
\begin{tabular}{lcccccc}
\toprule
\textbf{weight} & 0 & 0.01 & 0.1 & 0.5 & 1.0 & 2.0 \\
\midrule
$\lambda_{rel}$ 
& 65.44 & 67.61 & 69.31 & 72.44 & 73.15 & 72.41 \\
$\lambda_{attn}$
& 66.79 & 70.98 & 73.15 & 72.54 & 72.42 & -  \\
\bottomrule
\end{tabular}
\caption{Loss-weight sensitivity with eight latent reasoning steps at inference. Scores are mean accuracy (\%) over $V^*$ Bench, MMVP, and BLINK. We fix $\lambda_{\mathrm{ans}}=1.0$ and vary one auxiliary weight while keeping the other at its default value ($\lambda_{\mathrm{rel}}=1.0$, $\lambda_{\mathrm{attn}}=0.1$).}
\label{tab:ablation_lambda}
\end{table}

\section{Dataset Construction}
\label{app:dataset}

\subsection{Data Sources and Construction}
\label{app:data_sources}

\dataset contains 351,139 examples. Table~\ref{tab:dataset_source_counts} gives the exact source counts and proportions underlying the rounded counts in Table~\ref{tab:dataset}. We use fixed source quotas rather than implicit source rebalancing. Examples are selected after source-specific filtering with a fixed random seed and then shuffled.

\begin{table}[t]
\centering
\small
\setlength{\tabcolsep}{4pt}
\renewcommand{\arraystretch}{1.08}
\begin{tabular}{@{}lrr@{}}
\toprule
\textbf{Source} & \textbf{Examples} & \textbf{Share (\%)} \\
\midrule
GQA & 245,000 & 69.8 \\
OpenImages & 50,000 & 14.2 \\
CLEVR & 25,000 & 7.1 \\
Visual Genome & 15,139 & 4.3 \\
PSG & 6,000 & 1.7 \\
mGrounding & 5,000 & 1.4 \\
VSR & 5,000 & 1.4 \\
\midrule
\textbf{Total} & \textbf{351,139} & \textbf{100.0} \\
\bottomrule
\end{tabular}
\caption{Exact source composition of \dataset after source-specific filtering and sampling. Percentages are rounded.}
\label{tab:dataset_source_counts}
\end{table}

As described in Section~\ref{sec:trainingdata}, each sample contains an image, a question-answer pair, a scene graph, and bounding boxes for the objects involved in the scene graph or object relations. Most source datasets provide images, scene graphs, and object bounding boxes directly. VSR~\citep{liu2023visualsr} does not include bounding boxes, but it is built on COCO~\citep{lin2014microsoft}, which provides object bounding box annotations for the corresponding images.

Of the full dataset, 270,000 examples (76.9\%) from GQA and CLEVR natively provide the image, question-answer pair, relations, object identities, and boxes, requiring no additional annotation. The remaining 81,139 examples (23.1\%) use generated relation-focused QA. For this partition, only the questions and answers are generated; relations, object identities, and boxes remain source-provided. We use GPT-5.4 with the system prompt in Figure~\ref{fig:system_prompt}. Validation and human-audit procedures are detailed in Appendix~\ref{app:qa_quality}.

\begin{figure*}[p]
\centering
\begin{tcblisting}{
    enhanced,
    width=0.95\textwidth,
    colback=white,
    colframe=gray!35,
    coltitle=white,
    colbacktitle=gray!45,
    title={\textbf{System Prompt for Training Data Construction}},
    fonttitle=\bfseries\normalsize,
    boxrule=1.0pt,
    arc=0pt,
    outer arc=0pt,
    left=5pt,
    right=5pt,
    top=5pt,
    bottom=5pt,
    listing only,
    listing options={
        basicstyle=\ttfamily\fontsize{6.5pt}{7.2pt}\selectfont,
        breaklines=true,
        breakatwhitespace=true,
        columns=fullflexible,
        keepspaces=true,
        showstringspaces=false,
        aboveskip=0pt,
        belowskip=0pt
    }
}
You are a careful vision-language dataset annotator. Your task is to generate high-quality question-answer pairs from an image and its structured annotations.

Input:
You will receive:
1. image: the image to be annotated;
2. objects: a list of objects in the image. Each object contains:
   - object_id: a unique identifier;
   - name: the object category or noun phrase;
   - attributes: optional visual attributes;
   - bbox: [x_min, y_min, x_max, y_max];
3. relations: a list of scene-graph relations. Each relation contains:
   - subject_id;
   - subject_name;
   - relation;
   - object_id;
   - object_name;
4. target_relation_candidates: an optional subset of relations selected as candidate targets.

Task:
Generate one question-answer pair grounded in one or more scene-graph relations. If target_relation_candidates is provided and non-empty, select the target relation(s) only from this subset. Otherwise, select the target relation(s) from the full relations list.

Generation Requirements:
1. The question must be answerable from the image and the provided annotations.
2. The question must be grounded in explicit relation(s) from the scene graph, such as subject-relation-object, or in a directly annotated visual attribute of an object.
3. The answer must be short, accurate, and uniquely determined by the selected target relation(s) or attribute(s).
4. The target subject and object must be grounded by valid bounding boxes.
5. If multiple objects have the same name or category, the question must distinguish the intended object using visible attributes, spatial context, or relations to other objects.
6. The question should be natural and concise, but clarity and unambiguity are more important than linguistic variety.
7. Do not use external knowledge. Do not infer information that is not supported by the image and annotations.
8. Do not ask questions that require subjective judgment, uncertain visibility, fine-grained identity recognition, or private/sensitive attributes such as race, ethnicity, nationality, religion, gender identity, age, disability, health status, or socioeconomic status.
9. Do not generate unsafe, offensive, discriminatory, or inappropriate content.
10. Prefer rejecting the example over generating a question that is ambiguous, underspecified, weakly supported, or likely to have more than one valid answer.

Validation Before Output:
Before returning the result, verify that:
1. every target relation appears exactly in the provided relation list;
2. every supporting object appears in the provided object list and has a valid bounding box;
3. the question can be answered without using any information outside the image and annotations;
4. the answer is consistent with the selected target relation(s);
5. no other object or relation in the annotations would make a different answer equally valid;
6. the question wording does not introduce assumptions beyond the image and annotations.

Output Format:
Return only a JSON object. Do not include any additional text.

If a valid question-answer pair can be generated, return:
{
  "status": "ok",
  "question": "...",
  "answer": "...",
  "target_relations": [
    {
      "subject_id": "...",
      "subject_name": "...",
      "relation": "...",
      "object_id": "...",
      "object_name": "..."
    }
  ],
  "supporting_objects": [
    {
      "object_id": "...",
      "name": "...",
      "bbox": [x_min, y_min, x_max, y_max]
    }
  ],
  "rationale": "Briefly explain why the answer follows from the selected relation(s)."
}

If no safe, unambiguous, and well-grounded question can be generated, return:
{
  "status": "reject",
  "reason": "Briefly explain why the example is rejected."
}
\end{tcblisting}
\caption{System prompt used for constructing relation-grounded question-answer pairs.}
\label{fig:system_prompt}
\end{figure*}

\subsection{Deduplication and Split Separation}
\label{app:data_deduplication}

GQA uses Visual Genome images. We therefore collect the native VG image IDs in the selected GQA examples and remove all VG candidates with those IDs before sampling the VG quota. The retained GQA and VG subsets consequently have zero shared image IDs. We likewise make the VSR and PSG subsets disjoint using their common COCO image IDs and exclude COCO-derived records from mGrounding. These procedures remove the identified cross-source image overlaps from the final mixture.

We intentionally retain multiple questions about the same image when they query different facts or supervise different question-conditioned relations. These provide distinct supervision rather than duplicate records. We sample only from official training splits when available, and no official evaluation examples are used for training.

\subsection{Relation Vocabulary and Distribution}
\label{app:relation_distribution}

The relation vocabulary contains 2,111 labels and is fixed before training. Relation supervision uses standard instance-level cross-entropy without class weighting. Table~\ref{tab:relation_label_frequency} lists the five most frequent labels and representative labels from the long tail.

\begin{table}[t]
\centering
\small
\setlength{\tabcolsep}{4pt}
\renewcommand{\arraystretch}{1.08}
\begin{tabular}{@{}lr@{}}
\toprule
\textbf{Relation label} & \textbf{Frequency (\%)} \\
\midrule
\multicolumn{2}{@{}l}{\textit{Most frequent}} \\
\texttt{to\_the\_left\_of} & 15.3 \\
\texttt{to\_the\_right\_of} & 10.1 \\
\texttt{has\_attribute} & 4.1 \\
\texttt{on} & 2.0 \\
\texttt{wearing} & 2.0 \\
\midrule
\multicolumn{2}{@{}l}{\textit{Representative rare labels}} \\
\texttt{waving} & $<0.01$ \\
\texttt{cooking} & $<0.01$ \\
\texttt{feeding} & $<0.01$ \\
\texttt{pouring} & $<0.01$ \\
\texttt{surfing} & $<0.01$ \\
\bottomrule
\end{tabular}
\caption{Frequent and representative long-tail relation labels in \dataset. Frequencies are rounded; each listed rare label accounts for less than 0.01\%.}
\label{tab:relation_label_frequency}
\end{table}

\subsection{Generated-QA Validation and Human Audit}
\label{app:qa_quality}

Before inclusion, generated QA undergo deterministic checks for schema validity, object/relation references, box validity, answer consistency, and duplicates. These checks reject 4,327 candidates; all 81,139 generated-QA examples included in the training pool pass them.

The authors then audit 2,000 randomly selected unique examples from the generated-QA pool, inspecting the image, question-answer pair, source relation, object identities, and rendered boxes. An example is \textit{usable} if it has no correctness or grounding failure, allowing minor wording issues; \textit{strict-clean} permits no issue. A \textit{critical error} violates answerability, image/graph faithfulness, relation/object/box grounding, or the intended meaning. Table~\ref{tab:generated_qa_audit} summarizes the audit, including source-level quality and agreement between two independent human auditors on 400 examples.

\begin{table}[t]
\centering
\small
\setlength{\tabcolsep}{3pt}
\renewcommand{\arraystretch}{1.12}
\begin{tabularx}{\columnwidth}{@{}>{\raggedright\arraybackslash}p{0.38\columnwidth}>{\raggedright\arraybackslash}X@{}}
\toprule
\textbf{Audit measure} & \textbf{Result} \\
\midrule
Usable & 99.6\% (1,992/2,000); 95\% CI [99.2, 99.8]\% \\
Strict-clean & 98.1\% \\
Critical error & 0.4\% (8/2,000); 95\% CI [0.2, 0.8]\% \\
Lowest source-level usable rate & 97.5\% (200 audited examples) \\
Independent double audit & 98.0\% label agreement; Cohen's $\kappa=0.86$ on 400 examples \\
\bottomrule
\end{tabularx}
\caption{Human audit of generated QA before the follow-up ambiguity screening. Confidence intervals use the Wilson method; the lowest source-level usable rate is reported with its audit size.}
\label{tab:generated_qa_audit}
\end{table}

All eight critical errors are ambiguous generated questions without a unique, defensible answer; none involve incorrect source relations, object identities, or boxes. Following this audit, we screen the complete generated-QA partition for this failure mode and remove questions that do not admit a unique answer. After all validation and ambiguity screening, the final generated-QA partition contains 81,139 examples. The downstream contribution of including generated QA is evaluated in Appendix~\ref{app:generated_qa_contribution}.

\section{Baseline Details and Evaluation}
\label{app:baselines}

\subsection{Checkpoints and Result Provenance}
\label{app:checkpoints}

For locally evaluated open-source reasoning baselines, we use author-released checkpoints to preserve their respective published training procedures. Table~\ref{tab:baseline_checkpoints} lists their backbones and evaluated checkpoints. SLVR and HyLaR are included using author-reported results, as marked in Table~\ref{tab:main_results}. Except for LLaVA-OneVision and the proprietary GPT-4o model, all compared models use the Qwen2.5-VL-7B backbone.

\begin{table*}[t]
\centering
\small
\setlength{\tabcolsep}{3.5pt}
\renewcommand{\arraystretch}{1.05}
\begin{tabular}{@{}lll@{}}
\toprule
\textbf{Method} & \textbf{Backbone} & \textbf{Evaluated checkpoint} \\
\midrule
LLaVA-OneVision & SigLIP-SO400M + Qwen2-7B & \nolinkurl{lmms-lab/llava-onevision-qwen2-7b-ov} \\
GPT-4o & Proprietary & API model \\
Qwen2.5-VL-7B & Qwen2.5-VL-7B & \nolinkurl{Qwen/Qwen2.5-VL-7B-Instruct} \\
DeepEyes & Qwen2.5-VL-7B & \nolinkurl{ChenShawn/DeepEyes-7B} \\
PAPO & Qwen2.5-VL-7B & \nolinkurl{PAPOGalaxy/PAPO-G-H-Qwen2.5-VL-7B} \\
Vision-R1 & Qwen2.5-VL-7B & \nolinkurl{Osilly/Vision-R1-7B} \\
LVR & Qwen2.5-VL-7B & \nolinkurl{vincentleebang/LVR-7B} \\
MONET & Qwen2.5-VL-7B & \nolinkurl{NOVAglow646/Monet-7B} \\
LASER & Qwen2.5-VL-7B & \nolinkurl{wybb/laser-7b} \\
\method & Qwen2.5-VL-7B & Our trained checkpoint \\
\bottomrule
\end{tabular}
\caption{Backbones and evaluated checkpoints of the models compared in our experiments.}
\label{tab:baseline_checkpoints}
\end{table*}

\subsection{Baseline Descriptions}
\label{app:baseline_descriptions}

\paragraph{LLaVA-OneVision~\citep{li2024llava}.}
We include LLaVA-OneVision as a broadly trained open LVLM baseline. Unlike methods designed specifically for reasoning, LLaVA-OneVision is optimized for general visual instruction following across images, multi-image inputs, and videos. Its performance therefore reflects the capability of a strong general-purpose multimodal model.

\paragraph{Qwen2.5-VL-7B~\citep{bai2025qwen2}.}
Qwen2.5-VL-7B serves two roles in our experiments: it is both a strong open-source LVLM baseline and the initialization of \method. The model uses dynamic-resolution visual processing and has strong performance in visual tasks. Comparing with Qwen2.5-VL-7B directly measures the effect of adding our relation-grounded latent reasoning training.

\paragraph{DeepEyes~\citep{zheng2025deepeyes}.}
DeepEyes represents a line of work that improves reasoning by changing how the model interacts with visual evidence. Instead of relying only on the original visual input, it trains the model with reinforcement learning to perform image-in-the-loop reasoning, such as inspecting useful regions more carefully. This makes it a relevant comparison for methods that strengthen visual reasoning through additional perceptual actions.

\paragraph{PAPO~\citep{wang2025perception}.}
PAPO addresses the observation that multimodal RL can improve language-side reasoning while still leaving visual perception errors unresolved. It introduces perception-aware optimization so that policy learning is guided not only by answer correctness but also by whether the model preserves useful visual information. We compare with PAPO to evaluate against methods that explicitly target perception during reasoning.

\paragraph{Vision-R1~\citep{huang2025vision}.}
Vision-R1 brings R1-style post-training into multimodal reasoning. It first uses multimodal reasoning data for cold-start training and then applies reinforcement learning to elicit more deliberate reasoning behavior. It is included as a strong RL-enhanced LVLM baseline, complementary to our approach which improves reasoning through latent relation grounding rather than reward-based post-training.

\paragraph{LVR~\citep{li2025latent}.}
LVR is one of the closest baselines to our work because it also moves part of the reasoning process out of natural language. It trains autoregressive latent states to recover question-relevant visual tokens, allowing the model to reason in visual embedding space before generating the final answer. The key distinction is that LVR grounds latent states through visual reconstruction, while \method grounds them through question-relevant object relations.

\paragraph{MONET~\citep{wang2025monet}.}
MONET treats intermediate reasoning as continuous visual thoughts rather than decoded textual rationales. It uses staged training and latent-space policy optimization to make these continuous states useful for multimodal reasoning. We include MONET to compare with another recent framework that optimizes latent visual reasoning.

\paragraph{LASER~\citep{wang2026forest}.}
LASER improves latent visual reasoning through latent superposition, where a latent state is encouraged to carry information about a broader future reasoning window rather than a single immediate step. This design improves efficiency and preserves richer visual semantics during latent reasoning. 

\paragraph{SLVR~\citep{xu2026semantic}.}
SLVR enriches region-level visual latents with fine-grained semantics. It first aligns visual and semantic latent representations with region features, then applies Multi-query Group Relative Policy Optimization (M-GRPO) to encourage consistent representations across questions about the same region. We include SLVR as a baseline that adds explicit semantic supervision to latent visual reasoning.

\paragraph{HyLaR~\citep{cheng2026hylar}.}
HyLaR interleaves discrete text generation with continuous visual latent representations. It uses cold-start supervised fine-tuning followed by Decoupled Policy Optimization (DePO), which applies separate optimization constraints to text and latent actions. This provides a comparison with hybrid latent reasoning trained through reinforcement learning.

\section{Additional Results and Analysis}
\label{app:additional_results}

\subsection{Subset- and Task-Level Results}
\label{app:category_results}
\label{app:blink_results}

We provide task- and subset-level results for the benchmarks in Table~\ref{tab:main_results}, covering visual tasks, question categories, image resolutions, and difficulty levels. Table~\ref{tab:vstar_hrbench_subsets} reports the $V^*$ Bench and HRBench subsets, including author-reported results marked with $\dagger$. The SEED-Bench-2-Plus and HallusionBench tables report results for locally evaluated open-source models, using the same benchmark examples, prompts, image preprocessing, answer parsing, and scoring protocol as \method. Their checkpoints are listed in Appendix~\ref{app:checkpoints}.

We omit SLVR and HyLaR from the BLINK, SEED-Bench-2-Plus, and HallusionBench breakdowns because neither method had publicly available checkpoints at evaluation time, and their papers do not report the corresponding task- and category-level results.

\begin{figure}[htbp]
  \centering
  \includegraphics[width=\columnwidth]{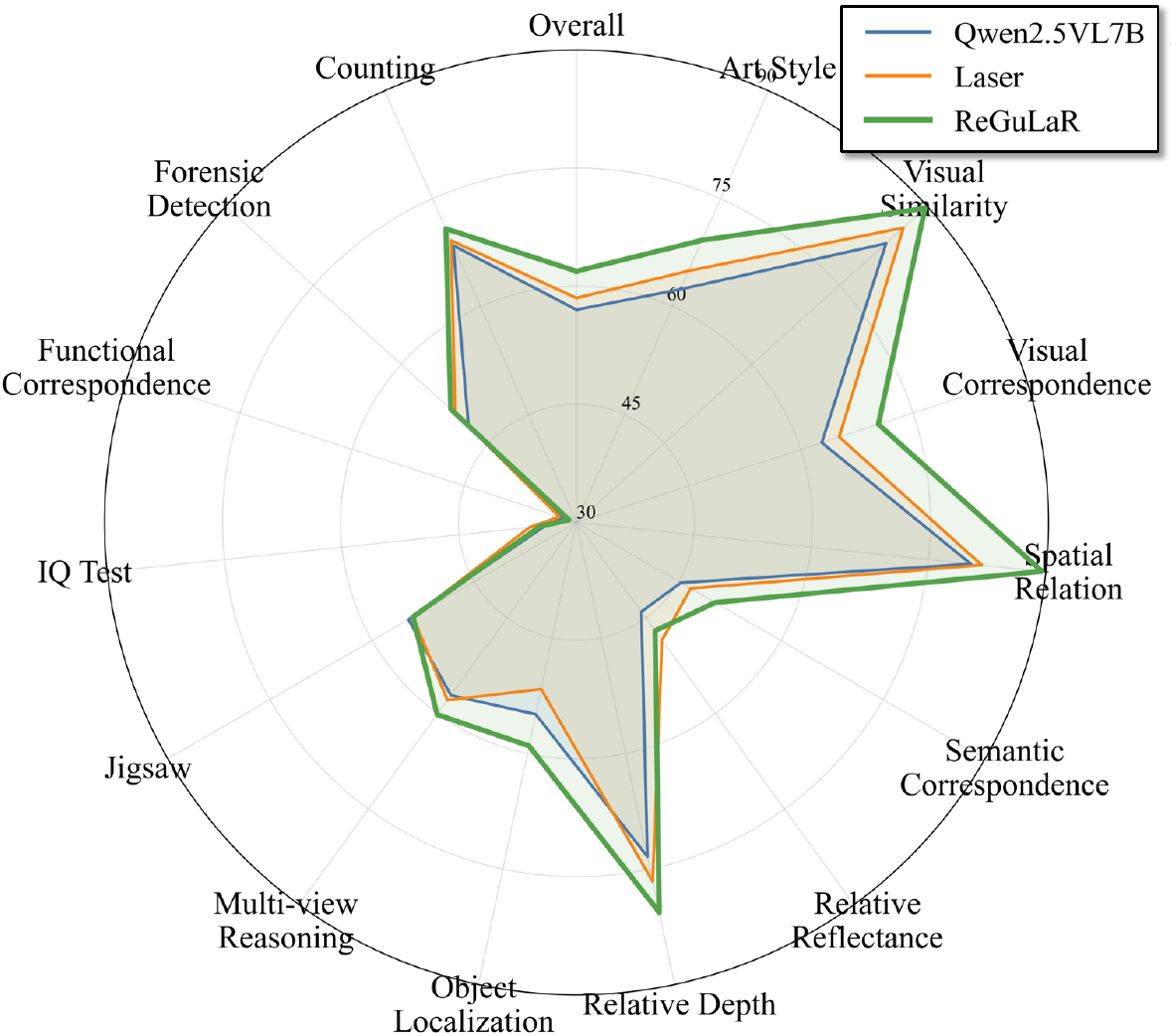}
  \caption{Task-level accuracy on the 14 BLINK tasks, comparing Qwen2.5-VL-7B, LASER, and \method.}
  \label{fig:blink}
\end{figure}

\paragraph{BLINK.}
BLINK comprises 14 diverse single-image and multi-image visual reasoning tasks. Figure~\ref{fig:blink} compares \method with its base model, Qwen2.5-VL-7B, and the latent reasoning baseline LASER. Among the three plotted models, \method achieves the highest accuracy on 10 of the 14 tasks. The gains are especially notable on tasks involving object attributes, spatial relations, and cross-image correspondence, suggesting that question-relevant relation grounding helps preserve fine-grained visual evidence across different reasoning settings.

\paragraph{$V^*$ Bench and HRBench.}
Table~\ref{tab:vstar_hrbench_subsets} separates $V^*$ Bench into Attribute and Spatial questions and HRBench into its 4K and 8K settings. The rows correspond to those in Table~\ref{tab:main_results}.

\begin{table*}[htbp]
\centering
\small
\setlength{\tabcolsep}{5pt}
\renewcommand{\arraystretch}{1.08}
\definecolor{opencolor}{RGB}{215,209,224}
\definecolor{closecolor}{RGB}{223,233,245}
\definecolor{greenbg}{RGB}{221,233,188}
\begin{tabular}{lcccc}
\toprule
\multirow{2}{*}{\textbf{Models}}
& \multicolumn{2}{c}{\textbf{V*}}
& \multicolumn{2}{c}{\textbf{HRBench}} \\
\cmidrule(lr){2-3}
\cmidrule(lr){4-5}
& Attribute & Spatial & 4K & 8K \\
\midrule
\rowcolor{greenbg}
\multicolumn{5}{c}{\textbf{\textit{General-purpose LVLMs}}} \\
GPT-4o
& 70.43 & 56.58 & 59.00 & 55.50 \\
LLaVA-OneVision
& 76.52 & 67.11 & 63.88 & 55.25 \\
Qwen2.5-VL-7B
& 79.13 & 72.37 & 68.00 & 61.75 \\
\midrule
\rowcolor{closecolor}
\multicolumn{5}{c}{\textbf{\textit{RL-based reasoning LVLMs}}} \\
DeepEyes
& 80.00 & 75.00 & 69.25 & 62.00 \\
PAPO
& 25.22 & 52.63 & - & - \\
Vision-R1
& 78.25 & \underline{78.95} & 66.63 & 60.25 \\
\midrule
\rowcolor{opencolor}
\multicolumn{5}{c}{\textbf{\textit{Latent reasoning LVLMs}}} \\
LVR
& 80.87 & 76.32 & 51.13 & 50.13 \\
MONET
& 81.74 & 76.32 & 67.37 & 60.88 \\
LASER
& 81.74 & 77.63 & 70.25 & 61.38 \\
SLVR-7B (Q1)$^{\dagger}$
& - & - & 70.40 & 62.50 \\
HyLaR-7B$^{\dagger}$
& \underline{82.61} & \textbf{85.53} & \textbf{75.00} & \textbf{70.50} \\
\midrule
\textbf{\method-7B}
& \textbf{83.48} & \textbf{85.53} & \underline{73.63} & \underline{66.25} \\
\addlinespace[2pt]
\rowcolor{black!5}
{\footnotesize\textit{vs Qwen2.5-VL-7B}}
& \gaincell{4.35} & \gaincell{13.16} & \gaincell{5.63} & \gaincell{4.50} \\
\bottomrule
\end{tabular}
\caption{Subset-level accuracy (\%) on $V^*$ Bench and HRBench. $\dagger$ marks author-reported results; SLVR uses SV-QA Q1. Dashes denote unreported values. Bold and underlined numbers denote the best and second-best displayed values, respectively. The shaded final row reports gains over Qwen2.5-VL-7B in percentage points.}
\label{tab:vstar_hrbench_subsets}
\end{table*}

\paragraph{SEED-Bench-2-Plus.}
The 2,277 public questions comprise 810 Chart, 807 Map, and 660 Web questions. Table~\ref{tab:seed_category_results} reports per-question accuracy for each category. \method improves over Qwen2.5-VL-7B by 4.81, 7.18, and 5.00 percentage points on Chart, Map, and Web, respectively. It also exceeds LASER by 1.11, 0.87, and 0.91 points, showing that the overall improvement extends across all three categories.

\begin{table}[htbp]
\centering
\small
\setlength{\tabcolsep}{4pt}
\renewcommand{\arraystretch}{1.05}
\begin{tabular}{@{}lccc@{}}
\toprule
\textbf{Method} & \textbf{Chart} & \textbf{Map} & \textbf{Web} \\
\midrule
LLaVA-OneVision & 55.06 & 55.02 & 76.36 \\
Qwen2.5-VL-7B & 63.95 & 54.28 & 80.45 \\
DeepEyes & 66.91 & 59.97 & 82.88 \\
PAPO & 51.98 & 54.65 & 56.06 \\
Vision-R1 & 67.04 & 59.73 & 82.57 \\
LVR & 42.96 & 55.02 & 43.48 \\
MONET & 62.10 & 55.89 & 82.73 \\
LASER & 67.65 & 60.59 & 84.54 \\
\midrule
\method & \textbf{68.76} & \textbf{61.46} & \textbf{85.45} \\
\bottomrule
\end{tabular}
\caption{Category-level accuracy (\%) on SEED-Bench-2-Plus: Chart ($n=810$), Map ($n=807$), and Web ($n=660$). Bold denotes the best score among the displayed models in each category.}
\label{tab:seed_category_results}
\end{table}

\paragraph{HallusionBench.}
Table~\ref{tab:hallusion_difficulty_results} reports per-question accuracy on 625 Easy and 504 Hard questions. \method improves over Qwen2.5-VL-7B by 8.96 and 11.50 percentage points on Easy and Hard questions, respectively. It achieves the highest Hard accuracy among the models in the table and ties with LASER for the highest Easy accuracy.

\begin{table}[htbp]
\centering
\small
\setlength{\tabcolsep}{5pt}
\renewcommand{\arraystretch}{1.05}
\begin{tabular}{@{}lcc@{}}
\toprule
\textbf{Method} & \textbf{Easy} & \textbf{Hard} \\
\midrule
LLaVA-OneVision & 60.32 & 40.08 \\
Qwen2.5-VL-7B & 75.52 & 34.13 \\
DeepEyes & 77.92 & 42.06 \\
PAPO & 70.88 & 37.70 \\
Vision-R1 & 79.20 & 43.05 \\
LVR & 82.56 & 45.43 \\
MONET & 69.28 & 37.10 \\
LASER & \textbf{84.48} & 45.43 \\
\midrule
\method & \textbf{84.48} & \textbf{45.63} \\
\bottomrule
\end{tabular}
\caption{Difficulty-level per-question accuracy (\%) on HallusionBench: Easy ($n=625$) and Hard ($n=504$). Bold denotes the best score among the displayed models in each difficulty group.}
\label{tab:hallusion_difficulty_results}
\end{table}

\subsection{Limited-Data Training}
\label{app:data_efficiency}

\begin{figure}[htbp]
\centering
\includegraphics[width=\columnwidth]{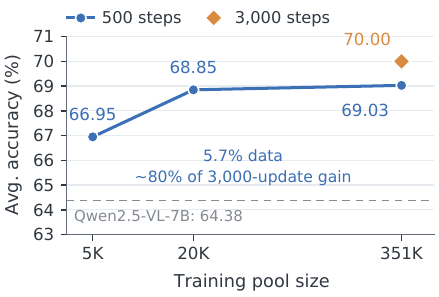}
\caption{Data efficiency of \method. Blue circles use 500 training updates with 5K, 20K, and 351K training pools; the orange diamond uses the 351K pool with 3,000 updates. The dashed line denotes pretrained Qwen2.5-VL-7B. Accuracy is averaged over all seven benchmarks in Table~\ref{tab:main_results}, with eight latent reasoning steps for trained variants. The logarithmic horizontal axis shows the available training pool, not the number of distinct examples seen during optimization. The gain annotation is relative to the pretrained baseline.}
\label{fig:data_efficiency}
\end{figure}

We assess data efficiency using random 5K and 20K subsets of \dataset alongside the full 351K training pool. Each pool is used to train \method for 500 optimization updates; a model trained on the full pool for 3,000 updates provides an additional reference. Figure~\ref{fig:data_efficiency} reports the arithmetic mean of overall accuracies across all seven benchmarks in Table~\ref{tab:main_results}, including MMStar. All trained variants use eight latent reasoning steps at inference. Here, updates refer to training optimization steps rather than latent reasoning steps at inference. A fixed update budget does not imply that every example in the training pool is visited.

With 500 updates, the 5K and 20K pools reach 66.95\% and 68.85\% average accuracy, exceeding the pretrained Qwen2.5-VL-7B baseline (64.38\%) by 2.57 and 4.47 percentage points, respectively. The 20K pool contains only 5.7\% of the full dataset and achieves about 80\% of the improvement of the 351K/3,000-update reference. At the same 500-update budget, increasing the pool from 20K to 351K raises accuracy by only 0.18 points to 69.03\%. Increasing training on the full pool from 500 to 3,000 updates further improves accuracy by 0.97 points to 70.00\%. These results show that relation-grounded training provides substantial gains with limited annotated data, while a larger training pool and longer optimization yield further improvements.

\subsection{Contribution of Generated QA}
\label{app:generated_qa_contribution}

We compare training on the 270K native-QA partition from GQA and CLEVR with training on the full \dataset, which additionally includes the 81,139 generated-QA examples. Both variants initialize from Qwen2.5-VL-7B and use the same 3,000-update training recipe. This ablation uses eight latent reasoning steps at inference for both variants. Table~\ref{tab:generated_qa_contribution} reports accuracy on $V^*$ Bench, MMVP, and BLINK, with Avg. computed over these three benchmarks.

\begin{table}[htbp]
\centering
\small
\setlength{\tabcolsep}{1.5pt}
\renewcommand{\arraystretch}{1.08}
\begin{tabular}{@{}lccccc@{}}
\toprule
\textbf{Training data} & \textbf{V*} & \textbf{MMVP} & \textbf{BLINK} & \textbf{Avg.} & \textbf{Retained} \\
\midrule
Native QA (270K) & 82.20 & \textbf{73.67} & 61.02 & 72.30 & 83.7\% \\
Full (351K) & \textbf{84.29} & \textbf{73.67} & \textbf{61.49} & \textbf{73.15} & 100.0\% \\
\bottomrule
\end{tabular}
\caption{Contribution of generated QA with matched initialization, 3,000 training updates, and eight latent reasoning steps at inference. Accuracy (\%) is reported on three benchmarks. Retained is the fraction of the full model's average gain over the pretrained backbone (67.93\%) retained by each variant, computed before rounding the averages.}
\label{tab:generated_qa_contribution}
\end{table}

Native-only training reaches 72.30\% average accuracy and retains 83.7\% of the full model's gain over the pretrained backbone. Including the generated-QA partition raises $V^*$ Bench and BLINK accuracy by 2.09 and 0.47 percentage points, respectively, while MMVP remains unchanged, yielding a 0.85-point average gain. Thus, native supervision accounts for most of the improvement, with additional gains when the audited generated-QA sources are included.

\section{Use of AI Assistants}
\label{app:ai_assistants}

We used AI assistants to support writing, coding, and data annotation. For writing, AI assistants were used to improve grammar and clarity, while all final text was reviewed by the authors. For coding, AI assistants helped with debugging and inspecting implementation details, with all code changes manually verified before use. AI assistants were also used in the data annotation pipeline; generation and quality-control procedures are detailed in Appendices~\ref{app:data_sources} and~\ref{app:qa_quality}.

\section{Potential Risks}
\label{app:risks}

\method is designed for research on vision-language reasoning and should not be used as a standalone system for high-stakes decision making. Like other LVLMs, it may produce incorrect or overconfident answers. \dataset is derived from existing public vision-language datasets and model-assisted annotation. It may therefore inherit biases, noise, or potentially sensitive visual content from the original data sources. We follow the licenses and terms of use of the source datasets, do not add personal metadata or identity labels, and manually inspect sampled generated annotations to reduce ambiguous or inappropriate question-answer pairs.

\section{Artifact Licenses}
\label{app:artifact_license}

\dataset is constructed from existing public vision-language datasets. We use these source datasets only for research purposes and follow their original licenses and terms of use. The derived dataset will be released subject to the licenses and redistribution constraints of the original data sources. The backbone and baseline models are used under their respective model licenses or terms of use. Our code will be released upon acceptance with a license specified in the release repository.

\section{Personally Identifying Information and Offensive Content.}
Our training data is derived from existing public vision-language datasets and does not introduce new personally identifying information beyond the source datasets. Since some source images may contain people, faces, scene text, or other potentially identifying visual content, we follow the access and redistribution terms of the original datasets. We do not collect additional personal information. We also manually inspect generated annotations to reduce ambiguous, offensive, or inappropriate question-answer pairs.

\end{document}